\documentclass[journal]{IEEEtran}
\ifCLASSINFOpdf
\fi

\ifCLASSOPTIONcompsoc
\usepackage[nocompress]{cite}
\else
\usepackage{cite}
\fi

\ifCLASSINFOpdf
   \usepackage[pdftex]{graphicx}
   \graphicspath{{../ImgJPG/}}
   \DeclareGraphicsExtensions{.pdf,.jpg,.png}
\else
  \usepackage[dvips]{graphicx}
  \graphicspath{{./Images/}}
  \DeclareGraphicsExtensions{.eps}
\fi

\usepackage{float}

\usepackage{amsmath}
\usepackage{algorithmicx}
\usepackage{algpseudocode}

\ifCLASSOPTIONcompsoc
  \usepackage[caption=false,font=normalsize,labelfont=sf,textfont=sf]{subfig}
\else
 \usepackage[caption=false,font=footnotesize]{subfig}
\fi
\usepackage{fixltx2e}
 \usepackage{dblfloatfix}

\usepackage{slashbox}
\usepackage{multirow}
\usepackage{booktabs}
\usepackage{color}
\usepackage{tablefootnote}
\usepackage{threeparttable}

\usepackage{url}

\begin{document}
%
\title{Object Detection by Spatio-Temporal Analysis \\ and Tracking of the Detected Objects \\ in a Video with Variable Background}
%
%
%

\author{Kumar~S.~Ray, 
        Vijayan~K.~Asari,~\IEEEmembership{Sr.,~Member,~IEEE,}
        and~Soma~Chakraborty
\thanks{This work was a part of collaborative research between Indian Statistical Institute and University of Dayton.}
\thanks{Kumar S. Ray is with Indian Statistical Institute, 203 B.T.Road, Kolkata-108, India 
					(e-mail: ksray@ isical.ac.in).}
\thanks{Vijayan K. Asari is with University of Dayton, 300 Collage park, Dayton, OH 45469-0232, USA 
					(e-mail: vasari1@udayton.edu).}
\thanks{Soma Chakraborty is with Indian Statistical Institute, 203 B.T.Road, Kolkata-108, India 
					(e-mail: soma.gchakraborty@gmail.com).}}

\maketitle

\begin{abstract}
In this paper we propose a novel approach for detecting and tracking objects in videos with variable background i.e. videos captured by moving cameras without any additional sensor. The performance of tracking in videos with variable background depends on the successful detection of an object in variable background. The most attractive feature of detecting an object in variable background is that it does not depend on any a priori information of the scene. In a video captured by a moving camera, both the background and foreground are changing in each frame of the image sequence. So for these videos, modeling a single background with traditional background modeling methods is infeasible and thus the detection of actual moving object in a variable background is a challenging task. To detect actual moving object in this work, spatio-temporal blobs have been generated in each frame by spatio-temporal analysis of the image sequence using a three-dimensional Gabor filter. Then individual blobs, which are parts of one object are merged using Minimum Spanning Tree to form the moving object in the variable background. The height, width and four-bin gray-value histogram of the object are calculated as its features and an object is tracked in each frame using these features to generate the trajectories of the object through the video sequence. In this work, problem of data association during tracking is solved by Linear Assignment Problem and occlusion is handled by the application of kalman filter. The major advantage of our method over most of the existing tracking algorithms is that, the proposed method does not require initialization in the first frame or training on sample data to perform. Performance of the algorithm has been tested on benchmark videos and very satisfactory result has been achieved. The performance of the algorithm is also comparable and superior with respect to some benchmark algorithms. 
\end{abstract}


%
\IEEEpeerreviewmaketitle

\section{Introduction}
%
%
%
%
\IEEEPARstart{O}{bject} tracking has emerged as a rigorous research topic due to increased demand of intelligent surveillance systems. But not only for surveillance, object detection and tracking are widely used in event classification, behavior understanding, crowd flow estimation, human-computer interaction and so on. Considerable progress is achieved in object tracking during the present decade and many benchmark algorithms have been established on object tracking \cite{cite3:Wu},\cite{cite4:Kristan}, behavior understanding \cite{cite19:Borges} etc. Still it remains a challenging problem due to the complexities in a video sequence like noise, unpredictable motion of objects, partial or full occlusion, illumination variation, background variation etc. When a video is captured by moving camera (hand-held or installed on a vehicle or on a rotating surface), both background and foreground features change their position in each frame. Thus, separation of background and foreground becomes highly complicated task as foreground extraction using traditional background subtraction or frame differencing does not apply for such videos. So far, some works have been reported to detect track object/s successfully in moving background. Most of the algorithms extract foreground by computing global motion and compensating it by processing feature points \cite{cite20:Arvanitidou},\cite{cite21:Choi},\cite{cite22:Lian},\cite{cite12:Zamalieva},\cite{cite29:Cui},\cite{cite23:Zhou},\cite{cite30:Lim},\cite{cite13:Ferone},\cite{cite24:Kim}, \cite{cite14:Hu1},\cite{cite5:Hu},\cite{cite31:Ghosh}. 

In our work, we have proposed a direct foreground extraction method by analyzing an input video spatio-temporally. In this approach, neither we need any apriori knowledge about the scene nor we take any assumption about the objects in the scene. The intuition behind the proposed method is that; rate of change of foreground region in consecutive frames is greater than the rate of change of background region, as motion in background region is only perceivable due to the translation or transformation of camera. But for foreground region, velocity of the region itself and the speed of camera are integrated. So, variation in foreground is more prominent than variation in background. If a block of few (say, 'n' number of) consecutive frames are analyzed at a time, then only significant spatial and temporal changes can be easily extracted suppressing the background variation. We have analyzed each spatio-temporal block of an input image sequence using a three-dimensional Gabor filter bank. Spatio-temporal blobs are then generated by applying selective average approach on the output of 3D Gabor filter bank. These blobs are either the whole moving object or parts of a moving object. So, Kruskal's Minimum Spanning Tree (KMST) is employed to merge the discrete blob-parts of an object. Then the object is represented by its height, width and gray-value histogram features. Based on these features, object is tracked using Linear Assignment Problem (LAP). We have also used Kalman filter to solve the problem of occlusion. Contributions of this paper are:
\begin{itemize}
\item The proposed algorithm applied a three-dimensional Gabor filter bank effectively to extract regions of motion in variable background in each input video frame without explicit feature point tracking or pseudo-motion compensation. The devised selective average approach efficiently generates spatio-temporal blobs by selecting pixels of high motion energy and suppressing non-informative or noisy pixels in regions of motion.
\item The proposed algorithm detects and tracks moving objects in videos captured by moving camera without any additional sensor or prior knowledge of environment or shape of objects. The major advantage of the method is that, it does not need initialization in first frame or training on sample data to perform. We have tested our algorithm on benchmark videos containing both variable and static background and achieved satisfactory performance. The algorithm has shown comparable and superior performance with respect to state-of-the-art methods \cite{cite5:Hu}, \cite{cite17:Zhang}, \cite{cite24:Kim}.  
\end{itemize}

The rest of the paper is organized as follows: in section (II) the related works on detection and tracking of moving objects in variable background are described, steps of proposed method are described in section (III), in section (IV) algorithm and complexity calculation is presented. Experimental results and analysis are depicted in section (V) followed by conclusion and future work in section (VI). 

\section{Related Works}

\subsection{On Spatial Analysis}
Most of the algorithms for detecting and tracking moving objects in variable background, classify foreground and background information by feature point or optical flow analysis with the aid of some geometric or probabilistic motion models. Some methods explicitly estimate and compensate global or camera motion to extract moving objects for tracking. Such as, in \cite{cite20:Arvanitidou} global motion is estimated using the Helmholtz Tradeoff Estimator and two motion models computed in prior frames. Fusing global motion compensated frames bidirectionally and applying thresholding and morphological operation moving object is segmented. In \cite{cite22:Lian}, Lian et al. estimated global motion by applying voting decision on a set of motion vectors determined by the edge features of objects or background. After removing edges with ego-motion, actual moving edges are enhanced by morphological operations. Zamalieva et al. estimated geometric transformations between two consecutive frames through dense motion fields using Geometric Robust Information Criterion (GRIC) \cite{cite12:Zamalieva}. Background/ foreground labels are obtained by combining motion, appearance, spatial and temporal cues in a maximum-a-posteri Markov Random Fields (MAP-MRF) optimization framework. Zhou et al. compensated the camera motion using a parametric motion model \cite{cite23:Zhou}. Moving objects are detected as outliers in the low-rank representation of vectorized video frames using Non-convex penalty and Markov Random Fields (MRFs). However, performance of this method in convergence to a local optimum depends on initialization of foreground support. Also it is not suitable for real-time object detection as it works in a batch mode. In \cite{cite13:Ferone}, adaptive neural self-organizing background model is generated to automatically adjust the background variations in each frame of video sequences captured by a pan-tilt-zoom (PTZ) camera. A registration mechanism is estimated in each frame to enable the neural background model to automatically compensate the ego-motion. In \cite{cite24:Kim}, camera motion is estimated by applying Lucas Kanade Tracker (LKT) to edges in current frame and background model. Moving objects are separated from background based on pixel-wise spatio-temporal distribution of Gaussian on non-panoramic adaptive background model. However, a PTZ camera can give maximum 360 degree view of a region as it's center is fixed. Thus it provides favorable condition for creating a background model. But video captured by a mobile camera captures a wide range of scene for which modeling a background is much more challenging. Authors of \cite{cite36:Yi} modeled a background by compensating camera motion using kanade Lucas Tracker (KLT) and applied dual-mode single Gaussian Model (SGM) with age to cope with motion compensation error.

Some methods simultaneously process foreground and background information using some motion models and detect moving objects for tracking without estimating camera motion explicitly. Like, in \cite{cite21:Choi}, authors simultaneously tracked multiple objects and estimated camera motion by finding the maximum-a-posteri (MAP) solution of the joint probability. Possible trajectories are sampled by reversible jump Markov chain Monte Carlo (RJ-MCMC) particle filtering. However, this method requires depth sensors which limits its application. In \cite{cite29:Cui}, authors decomposed a dense set of point trajectories into foreground and background using low rank and group sparsity based model. Classified trajectories are used to label foreground pixels in frame level. However, this method is prone to failure due to erroneous point tracking as it depends on tracking motion trajectories. In \cite{cite30:Lim}, temporal model propagation and spatial model composition are combined to generate foreground and background models and likelihood maps are computed based on these models. Then, graph-cut method is applied as energy minimization technique to the likelihood maps for segmentation. Authors of \cite{cite14:Hu1} and \cite{cite5:Hu} have extracted the feature points in the frames using standard feature point detection algorithms and classified them as foreground or background points by comparing optical flow features with multiple-view geometry. Foreground regions are obtained through image differencing and integrating classified foreground feature points. Moving object is detected using motion history and refinement schemes on foreground region. While \cite{cite14:Hu1} performs well in detecting slowly moving objects; \cite{cite5:Hu} provides better performance in detecting fast moving objects. In \cite{cite31:Ghosh} each video frame is segmented into a number of homogenous regions using fuzzy edge strength of each pixel in MRF modeling. Moving object is detected by calculating shift of centroids in successive frames and $\chi^2$-test-based local histogram matching of homogeneous regions. However, this approach has short-comings in presence of shadows or occlusion. Part-based deformable models are also applied to detect object in moving background scenario \cite{cite6:Hou}, \cite{cite32:Cho}. However, these methods need to detect actual moving object either by manual identification or by employing any off-the-shelf object detector. 

Object Tracking is also considered as multi-target association problem as in \cite{cite39:collins}, \cite{cite40:Shi}. In \cite{cite39:collins}, an iterative approximate solution is applied to a k-partite graph of observations from all the input frames. In \cite{cite40:Shi}, an $l_1$ tensor power iteration is introduced to solve the rank-1 tensor approximation problem where all trajectory candidates form a high-dimensional tensor.Challenge of abrupt motion handled in \cite{cite38:Lim} by a variant of traditional Particle Swarm Optimization (PSO) which self-tunes acceleration parameters by utilizing the averaged velocity information of the particles. In \cite{cite37:Kwon}, N-Fold Wang-Landau (NFWL)-based sampling method is used with a Markov Chain Monte Carlo (MCMC)-based tracking framework for both smooth and abrupt motion tracking.    

\subsection{On Spatio-temporal Analysis}
Spatio-temporal analysis is applied to detect object for tracking by exploiting statistical information on spatio-temporal correlation between object and its neighboring region \cite{cite17:Zhang}, by applying spatio-temporal Markov Random Field \cite{cite28:Khatoonabadi}, by training Hidden Markov Model \cite{cite33:Kratz}, Convolutional Neural Net \cite{cite8:Fragkiadaki} on spatio-temporal motion pattern or by using spatio-temporal tubes in unsupervised environment \cite{cite2:Xiao}. Object is also detected and tracked using spatio-temporal graph \cite{cite34:Sabirin}, space-time object contour \cite{cite9:Monma} and space-time deformable part-based models \cite{cite10:Fradi}. However, authors of \cite{cite17:Zhang} initialize target location in first frame manually or using some object detection algorithm and \cite{cite34:Sabirin},\cite{cite9:Monma},\cite{cite10:Fradi} are applicable only to static camera environment.
Similar such concept has also been exploited by other researchers on human action and gesture recognition. Such as, human action has been identified through learning spatio-temporal motion features \cite{cite1:Liu} or action is detected and described by Spatio-temporal interest points \cite{cite25:Xia}, Gaussian/Laplacian pyramid of  motion features \cite{cite16:Shao}, spatio-temporal segments of human body \cite{cite7:Harbi}, 3D covariance descriptors classified by weighted Riemannian locality preserving projection \cite{cite26:Sanin}, spatio-temporal bag of features to evaluate optical flow trajectories \cite{cite27:Wang}. In \cite{cite15:Tran}, events are detected by searching spatio-temporal paths.

\begin{figure*}[t]
\centering
\subfloat[]{\includegraphics[width=1in]{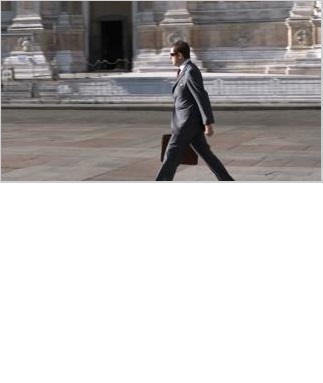}
\label{fig_1A}}
\hfil
\subfloat[]{\includegraphics[width=2.5in]{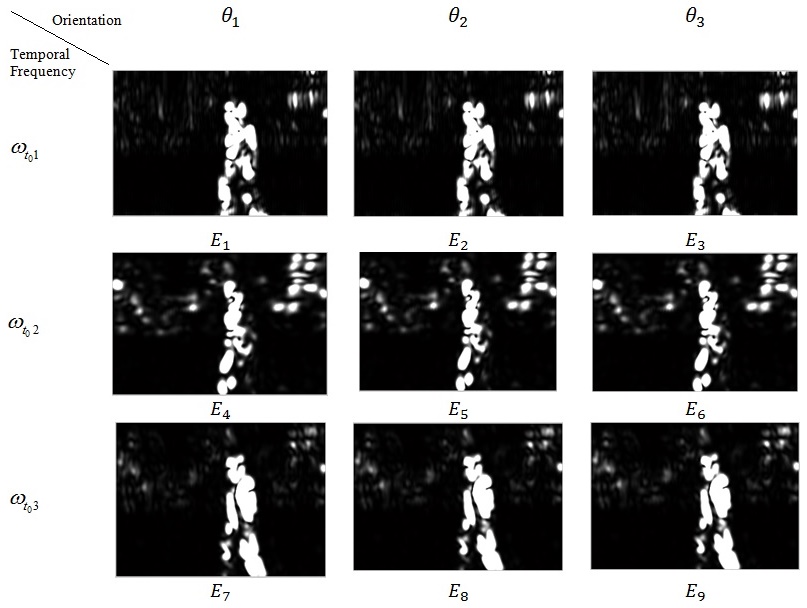}
\label{fig_1B}}
\hfil
\subfloat[]{\includegraphics[width=1in]{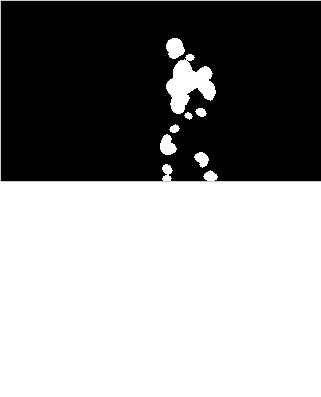}
\label{fig_1C}}
\caption{(a) Original frame of an image sequence. (b) Spatio-temporal energy responses. (c) Spatio-temporal blobs derived from Spatio-temporal energy responses.}
\label{fig_1}
\end{figure*}

\section{Proposed Method}

\subsection{Calculate spatio-temporal blobs}

Spatio-temporal blobs have been generated in three sequential steps- first, a spatio-temporal Gabor energy filter bank is created and spatio-temporal image blocks are devised from the input image sequence. In the next phase, each three dimensional image block is convolved with the three dimensional filter bank. In the third phase, spatio-temporal blobs are formed applying a selective average approach on the convolution results.
A three dimensional Gabor energy filter can be expressed as : 
\setlength{\arraycolsep}{0.0em}
\begin{eqnarray}
\label{eqn_1}
g(x,y,t,\omega,\sigma)&{}={}&\frac{1}{(2\pi)^{3/2}\sigma_x \sigma_y \sigma_t}\nonumber\\
  &&\cdot exp⁡(-\frac{1}{2} (\frac{x^2}{\sigma_x^2}+\frac{y^2}{\sigma_y^2}+\frac{t^2}{\sigma_t^2})) \nonumber\\
 &&\cdot exp(i(\omega_{x_0}x+\omega_{y_0}y+\omega_{t_0}t))
\end{eqnarray}
\setlength{\arraycolsep}{5pt}
where, $(\omega_{x_0},\omega_{y_0},\omega_{t_0})$ are the spatial and temporal center frequencies for which the filter will produce maximum spatio-temporal energy and $(\sigma_x,\sigma_y,\sigma_t)$ are the standard deviations of the three dimensional Gaussian envelope for the complex sinusoid of spatio-temporal Gabor filter; which determine the spatial and temporal extent of the kernel. $\omega_{x_0}$, $\omega_{y_0}$ are calculated using a base spatial frequency $(\omega)$ and spatial orientation $(\theta)$ as per following equations:
\begin{equation}
\label{eqn_2}
\omega_{x_0}=\omega\ast\cos(\theta) 
\end{equation}
\begin{equation}
\label{eqn_3}
\omega_{y_0}=\omega\ast⁡\sin(\theta)
\end{equation}
Various Spatio-temporal Gabor filters can be made by varying the $(\omega_{x_0},\omega_{y_0},\omega_{t_0})$ trio and a Spatio-temporal Gabor filter bank is created by combining those multiple three dimensional Gabor filters. In this work, we have parametrized Gabor filter bank by single base spatial frequency ($\omega$ $cycles/pixel$), three orientations $(\theta)$ and three center temporal frequencies ($\omega_{t_0}$ $cycles/frame$). The center spatial frequencies $(\omega_{x_0},\omega_{y_0})$ in horizontal and vertical directions have been calculated as per (\ref{eqn_2}) and (\ref{eqn_3}) using base spatial frequency and each of the orientations. Thus, we have total nine three dimensional Gabor filters for each $(\omega_{x_0},\omega_{y_0},\omega_{t_0})$ combinations.

In this work, a spatio-temporal image block for a frame is created by combining $(n-1)$ consecutive previous frames of the current frame and the current frame itself. Such as, spatio-temporal block of $p^{th}$ frame of the image sequence contains total 'n' no. of consecutive frames - $(p-(n-1))^{th},(p-(n-2))^{th}...p^{th}$; where $n<<M$ for M = length of the input image sequence. The value of 'n' depends on the temporal extent of the spatio-temporal Gabor filter i.e. n = temporal extent of the kernel.

Next, a spatio-temporal image block is convolved with the spatio-temporal Gabor energy filter bank to extract the prominent spatio-temporal variations corresponding to the moving object\/s. For the ease of computation and better performance the convolution is done by separable components of the three dimensional Gabor filter bank- odd-phase and even-phase Gabor filters with identical parameters \cite{cite35:Heeger}:
\setlength{\arraycolsep}{0.0em}
\begin{eqnarray}
\label{eqn_4}
g_{odd} (x,y,t,\omega,\sigma)&{}={}&\frac{1}{(2\pi)^{3/2}\sigma_x \sigma_y \sigma_t} \nonumber\\
                           &&\cdot exp⁡(-\frac{1}{2} (\frac{x^2}{\sigma_x^2}+\frac{y^2}{\sigma_y^2}+\frac{t^2}{\sigma_t^2}))\nonumber\\
                           &&\cdot \sin⁡(2\pi\omega_{x_0}x+2\pi\omega_{y_0}y+ 2\pi\omega_{t_0}t)
\end{eqnarray}
\setlength{\arraycolsep}{5pt}																						
 \setlength{\arraycolsep}{0.0em}
\begin{eqnarray}
\label{eqn_5}
g_{even} (x,y,t,\omega,\sigma)&{}={}&\frac{1}{(2\pi)^{3/2}\sigma_x \sigma_y \sigma_t}\nonumber\\
                           &&\cdot exp⁡(-\frac{1}{2} (\frac{x^2}{\sigma_x^2}+\frac{y^2}{\sigma_y^2}+\frac{t^2}{\sigma_t^2}))\nonumber\\
                           &&\cdot\cos⁡(2\pi\omega_{x_0}x+2\pi\omega_{y_0}y+ 2\pi\omega_{t_0}t)
\end{eqnarray}
\setlength{\arraycolsep}{5pt}                                                                                                 
The energy response of odd-phase and even-phase Gabor filter are:
\begin{equation}
\label{eqn_6}
E_{odd} (\omega_x,\omega_y,\omega_t )=I(x,y,t)*g_{odd} (x,y,t,\omega,\sigma)
\end{equation}
\begin{equation}
\label{eqn_7}
E_{even} (\omega_x,\omega_y,\omega_t )=I(x,y,t)*g_{even} (x,y,t,\omega,\sigma)
\end{equation}                                          
Spatio-temporal energy content of the current frame is calculated as the squared sum of odd and even-phase filter response:
\begin{equation}
\label{eqn_8}
E(\omega_x,\omega_y,\omega_t )=\{E_{odd} (\omega_x,\omega_y,\omega_t ) \}^2 +\{E_{even} (\omega_x,\omega_y,\omega_t )\}^2
\end{equation}                           
A filter with center frequencies $(\omega_{x_0},\omega_{y_0},\omega_{t_0})$ will give a larger output $E(\omega_x,\omega_y,\omega_t )$ when the convolved spatio-temporal signal has enough energy near the center frequencies. That is, a filter produces high responses for a pattern in a spatio-temporal image block which have similar frequencies as filter's center frequencies. 

In the present work, convolving the spatio-temporal image block with nine Gabor energy filters have produced nine such energy responses for the current frame. The energy responses- each of a size of the frame are denoted as $E_1.... E_9$ and are shown as in Fig. \ref{fig_1B}. Therefore each pixel in the current frame can be associated with an energy vector- $[e_1,e_2,e_3,e_4,e_5,e_6,e_7,e_8,e_9]$, where $e_1$ corresponds to $E_1$, $e_2$ corresponds to $E_2$ and so on for a particular pixel of each $E_n$. 

It is observed that, some pixels have high energy values in most of the fields of its energy vector whereas some pixels have lower energy values consistently. So, it can be said that pixels with lower energy vectors contain little or no spatio-temporal information for the current frame and are therefore discarded in further processing. Some pixels of the frame has inconsistent energy vector, i.e. few fields contain large value where most of the fields contain very low energy values. These pixels are considered as noise and are discarded. Selection of pixels of high energy vector and rejection of non-informative and noisy pixels to create the spatio-temporal blobs is described next.

Each field of an energy vector of a pixel is marked as accepted if its value is greater than or equal to the standard deviation of the corresponding energy response frame $E_n$, i.e.

\begin{equation}
\label{eqn_9}
{mark}_n = \begin{cases}
      e_n & ; e_n\geq\sigma(E_n) \\
      0   & ; \text{otherwise}
    \end{cases}
\end{equation}

For each pixel of the current frame, if number of accepted energy values is greater than no. of rejected energy values, then that pixel is assigned with the average of high energy values and pixel is set to zero otherwise, i.e.

\begin{equation}
\label{eqn_10}
P_{r,c} = \begin{cases}
      \frac{1}{accepted}\sum\limits_{i=1}^{n}{mark}_n & ; \text{accepted}>\text{rejected} \\
      0 & ; \text{accepted}<\text{rejected}
    \end{cases}
\end{equation}

where, $accepted=\#({mark}〗_n>0)$ and $rejected=\#({mark}_n=0)$. 
Thus, a single energy response frame (E) is created from nine energy responses of the spatio-temporal filters. 'E' contains the desired spatio-temporal blobs as foreground or the region of interest as depicted in Fig. \ref{fig_1C}.

\subsection{Formation of moving object from discrete spatio-temporal blobs}

\begin{table*}[t]
\renewcommand{\arraystretch}{1.3}
\caption{A Sample Weight Matrix}
\label{table_wm}
\centering
\begin{tabular}{c c c c c c c c}
\hline
 $$ & $\textbf{Node}_1$ & $\textbf{Node}_2$ & $\textbf{Node}_3$ & $\textbf{Node}_4$ & $\textbf{Node}_5$ & $\textbf{Node}_6$ & $\textbf{Node}_7$\\
\hline\hline
$\textbf{Node}_1$ & 0 & 15.65	& 72.56 & 98.60 &	73.79 &	87.21 & 126.02 \\
$\textbf{Node}_2$ & 15.65 &	0	& 65.19 &	84.40 &	61.98 & 75.27 &	119.60 \\
$\textbf{Node}_3$ & 72.56 & 65.19	& 0 &	55.08 &	24.73 & 30.59 &	54.45  \\
$\textbf{Node}_4$ & 98.59 & 84.40 & 55.08 & 0	& 32.28 &	25.17 &	80.28  \\ 
$\textbf{Node}_5$ & 73.79 & 61.98 & 24.73 & 32.28 & 0	& 13.41 & 67.53  \\ 
$\textbf{Node}_6$ & 87.20 & 75.27 & 30.59 & 25.17 & 13.41 & 0	& 60.60  \\
$\textbf{Node}_7$ & 126.01 & 119.60 & 54.45 & 80.28 & 67.53 & 60.60 & 0 \\ 
\hline
\end{tabular}
\end{table*}

\begin{figure}[t]
\centering
\subfloat[]{\includegraphics[width=2in]{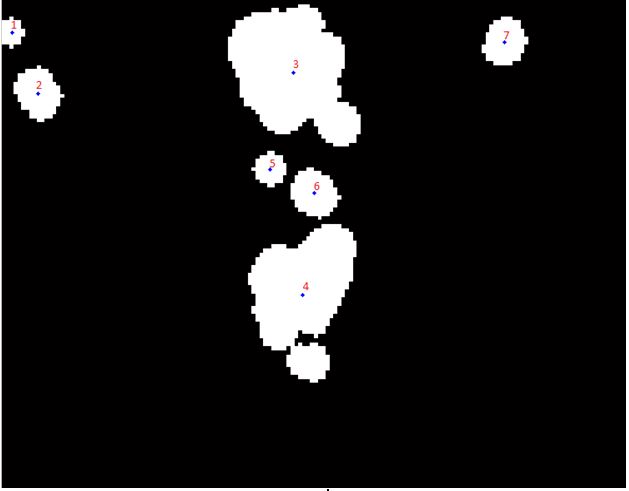}
\label{fig_2A}}
\hfil
\subfloat[]{\includegraphics[width=2in]{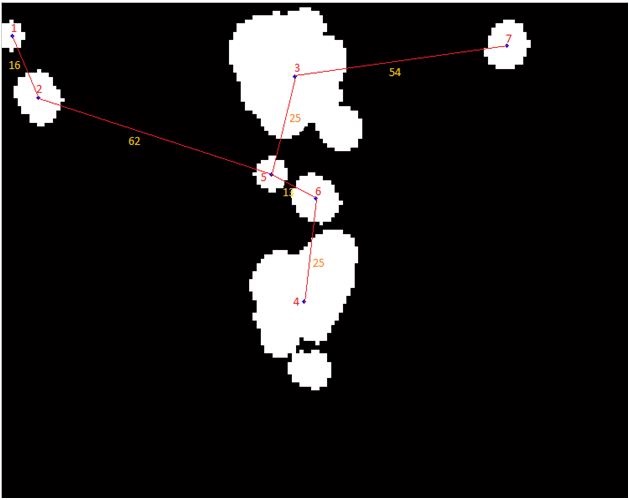}
\label{fig_2B}}
\hfil
\subfloat[]{\includegraphics[width=2in]{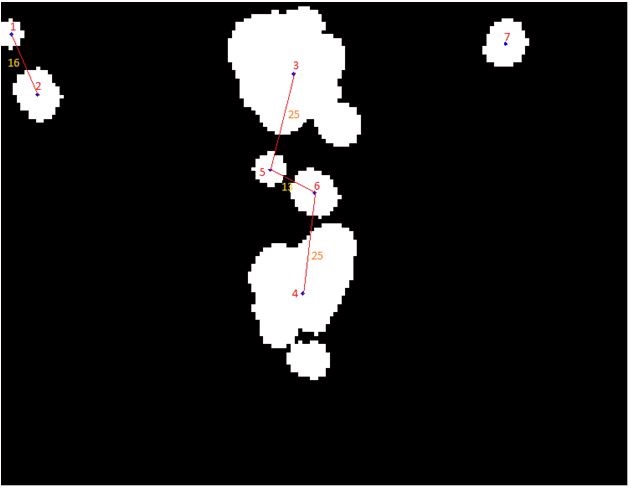}
\label{fig_2C}}
\caption{(a) Centroids of spatio-temporal blobs as nodes. (b) Minimum Spanning Tree for the centroids. (c) Sub-trees of  MST after thresholding.}
\label{fig_2}
\end{figure}

\begin{figure}[t]
\centering
\subfloat[]{\includegraphics[width=1in]{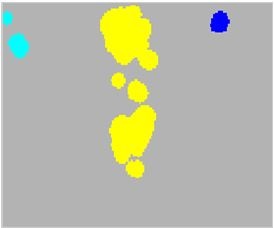}
\label{fig_3A}}
\hfil
\subfloat[]{\includegraphics[width=1in]{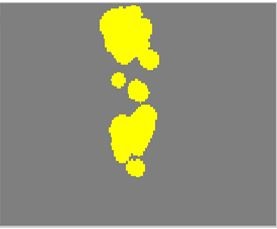}
\label{fig_3B}}
\caption{(a) Clustered Spatio-temporal blobs which are part of same object. (b) Significant cluster of Spatio-temporal blobs in a frame.}
\label{fig_3}
\end{figure}

Discrete spatio-temporal blobs may be parts of same object or the whole moving object in a frame. So in this phase we have processed blobs, so that fragments of same object are merged to mark the whole region containing the actual moving object. Almost-connected blobs are united on the basis of their proximity using Kruskal's Minimum Spanning Tree (MST) algorithm as follows:

Centroids of each spatio-temporal blob is considered as nodes to construct the MST as depicted in Fig. \ref{fig_2A}. A $u\times{u}$ weight matrix is formed where $u$ is the number of nodes (centroids) in each frame. Each cell of the matrix contains the Euclidean distance between each pair of the centroids. Each row and column of the matrix contains the euclidean distance between the node in that row/column and other nodes as shown in the Table \ref{table_wm}. Then Kruskal's MST algorithm is applied on the weight matrix and the minimum spanning tree for the frame is constructed as shown in the Fig. \ref{fig_2B}. Sum over average weight ($\overline{w}=mean(W)$) and standard deviation ($w_\sigma=\sigma(W)$) of the weight matrix is used as the threshold for comparing the edge-weight of the MST to determine the proximity of the blobs. Edges with weight greater than the threshold are removed from the tree and this leads to the formation of sub-trees which are groups of centroids. As it is apparent from Fig. \ref{fig_2B} that, egde(2,5) and edge(3,7) both are quite longer than other edges in the MST and have been removed from the tree by thresholding to produce sub-trees as displayed in Fig. \ref{fig_2C}. Blobs with centroids in the same group are clustered and labeled as one object as shown in the Fig. \ref{fig_3A}. After that, total area of each cluster is calculated and any cluster having an area less than one-third of the area of the largest cluster is removed. Cluster of significant size are considered as the actual moving object. Fig. \ref{fig_3B} depicts the frame after purging the very small cluster of blobs.

Each detected object is then represented by a feature vector consisted of- (1) Centroid of the object, (2) Height and width of the object and (3) four bin gray value histogram of the object. Higher number of bins for the histogram also have been tried on our test set of videos. But it is observed that, increasing or decreasing the number of histogram bins is not affecting distinguishability of individual object significantly. Also the number of bins is not affecting the performance of our tracking method. So, for the present set of input videos four-bin gray value histogram is considered as sufficient to represent an object.

\subsection{Tracking the object}
In this work, objects are tracked using Linear Assignment Problem (LAP). Each detected object of the current frame is compared and matched with an object in the previous frame. The problem of linear assignment is solved by calculating cost matrix for a pair of detected object and existing track in each frame except the first frame of the image sequence. In the first frame of an input sequence, for each detected object a track is created containing the features of the object described in the previous section. Now in the next frame a set of objects are detected among which some of the objects belong to the earlier frame and some are appearing for the first time. Here the question of linear assignment arises. 

\begin{figure}[t]
\centering
\includegraphics[width=3in]{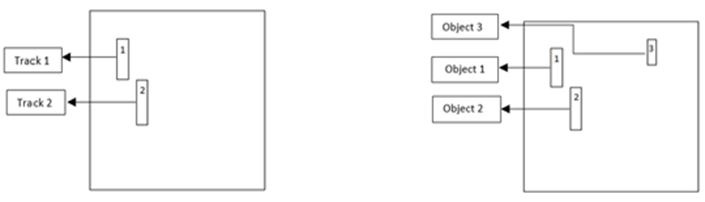}
\caption{Existing tracks of the previous frame (left) and detected objects in current frame (right).}
\label{fig_5}
\end{figure}

The cost of assigning each object from the detected set of objects in current frame to the existing track is calculated as:
\begin{equation} 
\label{eqn_20}
{cost}_{kl}=
 \begin{cases}
           \frac{1}{4}\sum\limits_{n=1}^{4} |{GH}_{k_n}-{GH}_{l_n}|, & \text{if d$\leq ({H}_k\&{W}_k)$}\\
					& \text{$\& (d_h\&d_w<5)$} \\
					
					\phi, & \text{Otherwise}
 \end{cases}	
\end{equation}
where, ${GH}_k$'s and ${GH}_l$'s are the four-bin gray value histograms of $k^{th}$ track and $l^{th}$ object respectively, ($d=\sqrt{({x_k}-{x_l})^2+({y_k}-{y_l})^2}$) is the distance between centroids of $k^{th}$ track and $l^{th}$ object, ($d_h=|{H}_k-{H_l}|$) and ($d_w=|{W}_k-{W}_l|$) are the absolute differences of heights (${H}_k, {H_l}$) and widths (${W}_k, {W}_l$) of the same track-object pair and $\phi$ is an arbitrary large value to indicate "no resemblance" between $k^{th}$ track and $l^{th}$ object. As in Fig. \ref{fig_5}, there are two existing tracks in the previous frame and three detected objects in the current frame. We compute the Euclidean distance between the centroid value stored in ${Track}_1$ and the centroid value of each object of the current frame (say, ${Object}_1...{Object}_N$). If distance between centroids is greater than the thresholds (height and width of the previously detected object) stored in ${Track}_1$ and if the difference of sizes of bounding boxes of ${Track}_1$ and any $n^{th}$ object is greater than 5, then we assign $-1$ value for the ObjectNo field and $\phi$ in Cost field of the cost matrix for the ${Track}_1-{Object}_n$ pair which indicates no resemblance; otherwise, ${Object}_n$ is considered as a promising candidate to be assigned to the ${Track}_1$. So, we proceed to compare between four-bin gray-value histograms of the ${Object}_n$ and the ${Track}_1$. The identification number (ID) of the Object is assigned for the ObjectNo field and average absolute differences of four-bin gray-value histograms is assigned to the Cost field of the cost matrix for the ${Track}_1-{Object}_n$ pair. This process is repeated for all detected objects for each of the existing tracks and cost matrix of each ${Track}_n-{Object}_n$ pair in the current frame is prepared. For example, Table \ref{table_cm} is a sample cost matrix for a video frame like Fig. \ref{fig_5}:

We update the Track for each frame in the following process:
An object is decided to be a perfect match with a track with minimum cost. So, for each track, cost matrix is searched for minimum cost and corresponding ObjectNo and the track is assigned to the object of ObjectNo  with minimum cost. For example, as ${Object}_1$, in Table \ref{table_cm} has minimum cost for ${Track}_1$; ${Object}_1$ will be assigned to ${Track}_1$. Thus all track data are updated with proper object data. The cost matrix is also updated by labeling the track as assigned and removing the whole object entry from the cost matrix. In any frame, there may be some objects which do not have any resemblance with any track, i.e. they may have $-1$ for ObjectNo and $\phi$ as cost for all the tracks in cost matrix; then these are new objects. We introduce n number of new tracks for n new objects and assign the new objects to the new tracks. Such as, ${Object}_3$ in Table \ref{table_cm} is a new object and will be assigned to new track ${Track}_3$. On the other hand, if any object has resemblance with one or more existing track/s, but is not assigned to any of the tracks; is considered as false detection or erroneous detection due to 'less than 50\% partial occlusion'. These objects are discarded. Also some objects detected in earlier frames (existing track in current frame) may not be matched with any object in current frame. These undetected objects are considered as either stopped or fully occluded. For these undetected objects kalman filter is applied to update its track information as described in the next section.

\begin{table}[t]
\renewcommand{\arraystretch}{1.3}
\caption{A Sample Cost Matrix For a Video Frame}
\label{table_cm}
\centering
\begin{tabular}{c c c}
\hline
$\textbf{Object}/\textbf{Track}$ & $\textbf{Track}_1$ & $\textbf{Track}_2$\\
\hline\hline
 {$\textbf{Object}_1$} & $Object no. = 1$ & $Object no. = 2$ \\
 & $Cost = 1.882$ & $Cost = 19.43$ \\
\hline
$\textbf{Object}_2$ & $Object no. = 1$ & $Object no. = 2$ \\
 & $Cost = 28.79$ & $Cost = 4.556$ \\
\hline
$\textbf{Object}_3$ & $Object no. = -1$ & $Object no. = -1$  \\
 & $Cost = \phi$ & $Cost = \phi$  \\
\hline
\end{tabular}
\end{table}

\subsection{Handling occlusion using Kalman filter}
To enable tracking of object during full occlusion, each newly detected object is assigned a Kalman filter. The state variable is parametrized as:
\begin{equation}
\label{eqn_11}
x_t=(r_t,c_t,v_{r_t},v_{c_t})^T
\end{equation}

where $(r_t,c_t) =$ Centroid of an object in $t^{th}$ frame and $(v_{r_t},v_{c_t}) =$ velocity of the object. For our present work velocity is assumed as constant. When an object is detected for the first time and assigned to a new track, corresponding Kalman filter of the object (as well as of the track) is initialized with the coordinates of its centroid i.e. $x_t=(r_{t_n},c_{t_n},0,0)^T$, where $(r_{t_n},c_{t_n}) =$ centroid of $n^{th}$ object in $t^{th}$ time frame and the velocity is initialized as zero. Other parameters of the Kalman filter are initialized as per standard practice: the state transition model is set as $A=[1010;0101;0010;0001]$. The measurement model which relates state variables to the measurement or output variables is set as $H=[10;00;01;00]$. State estimation error covariance matrix $P=100*I_{4\times4}$, system noise $Q=0.01*I_{4\times4}$ and measurement noise $R=I_{2\times2}$ or a unit matrix. When the track is reassigned in the next frame its corresponding Kalman filter is updated as the rules stated below:

A priori estimate of the state vector 'x' of any assigned track at time 't' using the state information of the previous frame (at time 't-1') is done first using (\ref{eqn_12})-
\begin{equation}
\label{eqn_12}
x_{t|t-1}=Ax_{t-1|t-1}
\end{equation}

A priori update of error covariance matrix P is done by (\ref{eqn_13})-
\begin{equation}
\label{eqn_13}
P_{t|t-1}=AP_{t-1|t-1}A^T+Q
\end{equation}

Then kalman gain $K_t$ is calculated to minimize a posteriori error covariance matrix P as per the rule in (\ref{eqn_14})-
\begin{equation}
\label{eqn_14}
K_t=P_{t|t-1}H^T/(HP_{t|t-1}H^T+R)
\end{equation}

A posteriori estimation of state 'x' at time 't' is calculated as (\ref{eqn_15}). It compares the actual measurement or the location of the object in current frame and the measurement from the priori state estimate (\ref{eqn_12}). Then using the difference between those two values and the kalman gain the rule determines the amount by which the priori state estimate is to be adjusted. 
\begin{equation}
\label{eqn_15}
x_{t|t}=x_{t|t-1}+K_t(y_t-Hx_{t|t-1})
\end{equation}

At last, the posteriori error covariance matrix is calculated by rule as per \ref{eqn_16}.
\begin{equation}
\label{eqn_16}
P_{t|t}=(I-K_tH)P_{t|t-1}
\end{equation}

If an existing track is not assigned an object in the current frame then it is assumed that, the object contained in the track is not detected due to occlusion. We use the Kalman filter prediction of the track to update the track record. That is, only (\ref{eqn_12}) and (\ref{eqn_13}) are calculated and the prior state estimate is used to update the centroid information for the unassigned track and all other information remain same as the previous record. Using this method in our method, we have achieved the continuation of tracking in presence of considerable occlusion.

\section{Algorithm And Performance}

\floatstyle{boxed} 
\restylefloat{figure}
\begin{figure*}[t]
\caption{Algorithm for object detection and tracking in complex and variable background}
\label{fig_mainAlg}
\begin{algorithmic}
\State \textbf{Algorithm:} Object detection and tracking in complex and variable background
\\ \State \textbf{Input:} $V$ = a video of size $m\times n\times T$;
 \State \textbf{Output:} $V_{annotated}$ = a video of same size of V marked with Object labels and bounding boxes;
\\ \State Let, G is a separable Three-dimensional Gabor filter of size $x\times y\times t$- $G_{odd}$ and $G_{even}$ are the odd-phase and even-phase components of G;
\\ \State Let, $I$ is a spatio-temporal image block of size $m\times n\times t$;
\\ \State \textit{Initialization:} 
\State Construct $G_{odd_1}...G_{odd_N}$  of n number of different temporal and spatial frequencies ;
 \State Construct $G_{even_1}...G_{even_N}$ of n number of different temporal and spatial frequencies ; 
 \begin{enumerate}
\For{$f=t$ to $T$}
\ \ \ \For{$index=1$ to $t$}
\ \ \ \ \ \ \State $I[m,n;index]=[V_{f-(t-1)},V_{f-(t-2)}...V_f]$;
\ \ \ \EndFor
 \State Calculate spatio-temporal blobs using (\ref{eqn_8}), (\ref{eqn_9}) and (\ref{eqn_10});
 \State Compute a weight matrix of the set of centroids of distinct spatio-temporal blobs;
 \State Generate Minimum Spanning Tree of centroids using Kruskal's algorithm;
 \State Generate clusters of centroids by thesholding each MST edge;
 \State [${Centroid}_n,{BB}_n,{GH}_n$] -feature vector of each the ${Object}_1,...,{Object}_n$;
 \State Let, ${Track}_{1\times N}$ is a set of tracks for N number of detected objects in current frame;
 \State Let, ${Kalman}_{1\times K}$ is the kalman prediction corresponding each track;
 \If {$f==t$ or starting operating frame of $V$ then}
\ \ \State Assign features of each detected object ${Object}_n$ to a track ${Track}_n$ ;
\ \ \State Initiate ${Kalman}_n$ for each ${Track}_n$ ; 
\Else
\ \  \State Calculate cost matrix using (\ref{eqn_20});
\ \ \State Call $Algorithm 1$ (see Appendix 1) using resultant cost matrix as argument
\EndIf
\ \ \State Mark centroid and identification number of each track on the current frame and display;
\EndFor
\end{enumerate}
\end{algorithmic}
\end{figure*}

The main steps of computation performed by the procedure are: 1) calculating spatio-temporal blobs, 2) clustering discrete blobs which are parts of same object by constructing a minimum spanning tree to estimate the proximity of the blobs, 3) creating object trajectories using LAP and 4) handling occlusion using Kalman Filter. The algorithm is depicted in Fig. \ref{fig_mainAlg}.

Convolving the image sequence with separable Gabor filter bank has the complexity of $O(kmnl)$ \cite{cite35:Heeger}, where \textit{k} is the size of convolution kernel, \textit{l} is the length of the image sequence and \textit{m}, \textit{n} are the height and width of the image sequence respectively. To calculate of spatio-temporal blobs nine energy frames are processed for each frame. That is nine values of each $(m\times n)$ pixels are compared with threshold and selected for averaging- total execution time of this process for each frame is $9\ast mn=O(mn)$. The cost of constructing Minimum Spanning Tree for each frame is $O(e\log v)$, where \textit{e} is the number of edges in the graph and \textit{v} is the number of vertices of the graph. The last part of the procedure is the Linear Assignment Problem (LAP). If there are \textit{x} existing tracks and \textit{y} detected objects in a frame, then complexity of computing cost matrix is $O(xy)$. But in average case, most of the track-object cost is not calculated if the centroid distance cost does not meet the threshold criteria. So, in average case the time of cost evaluation reduces drastically. The next part of LAP is assignment of objects to tracks. This phase has worst case time of $O(x^3)$ as per Jonker's  assignment algorithm, as most of the time the number of tracks are greater than the number of objects. Total execution time of the algorithm for each frame is $O(kmnl)+O(mn)+O(e\log v)+O(xy)+O(x^3)\approx O(kmnl)+O(x^3)$. 

\section{Experiments}

\floatstyle{plain}
\restylefloat{figure}
\begin{figure}[t]
\centering
\subfloat[]{\includegraphics[width=2.5in]{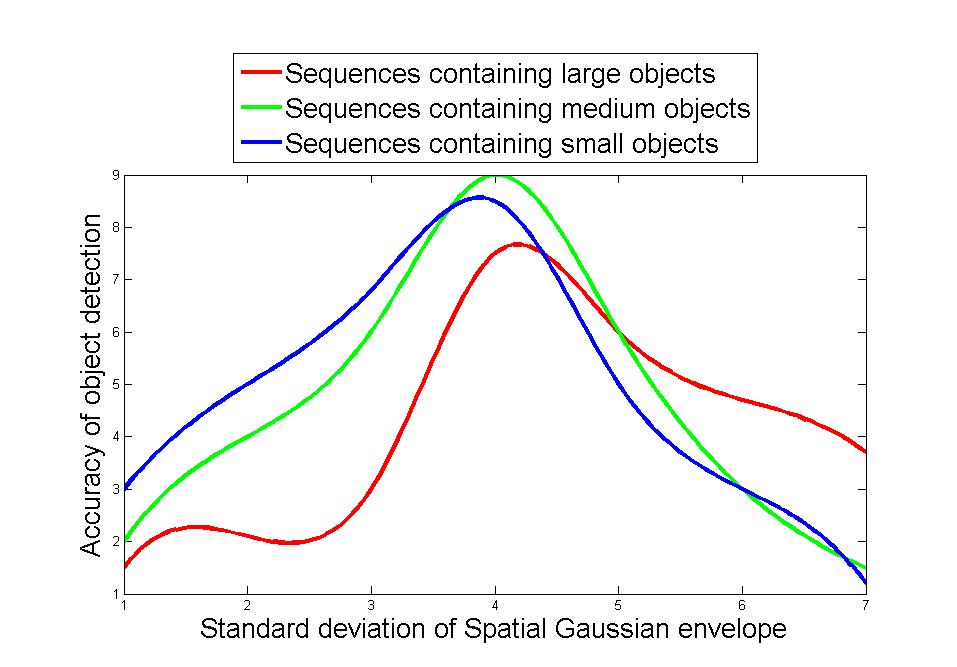}
\label{fig_6A}}
\hfil
\centering
\subfloat[]{\includegraphics[width=2.5in]{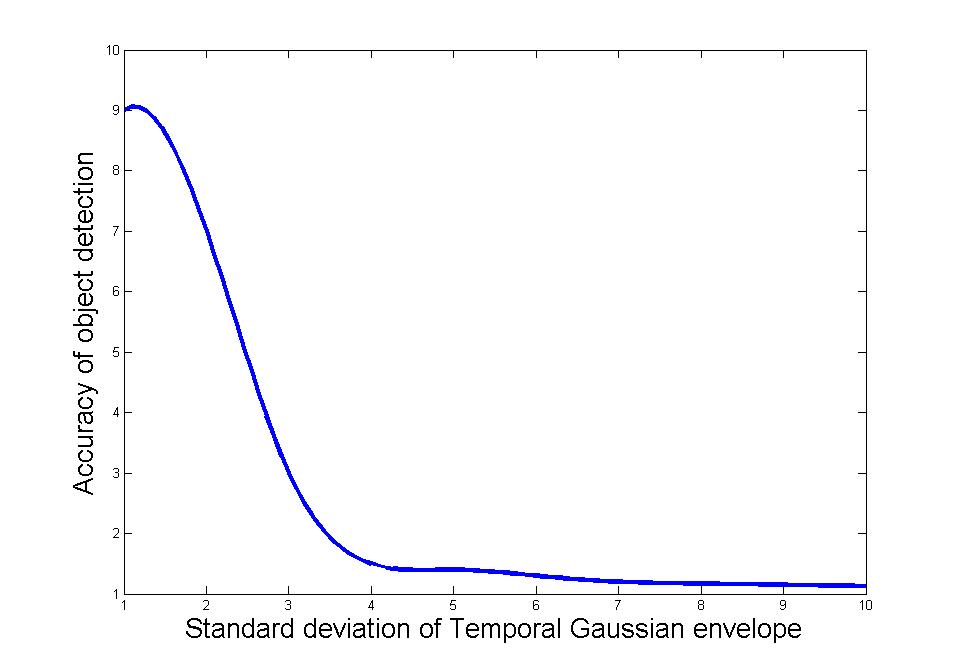}
\label{fig_6B}}
\caption{Effect of increase or decrease of standard deviations of Gaussian envelopes on the accuracy of object detection. (a) Accuracy of object detection Vs Standard deviation of Spatial Gaussian envelope. (b) Accuracy of object detection Vs Standard deviation of Temporal Gaussian envelope.}
\label{fig_6}
\end{figure}

\begin{figure*}
\centering
\subfloat{\includegraphics[width=2in]{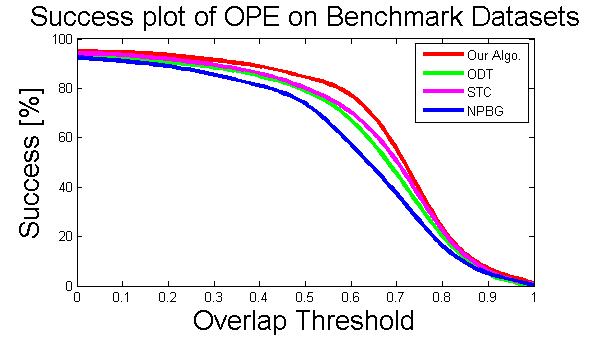}
\label{fig_go}}
\hfil
\subfloat{\includegraphics[width=2in]{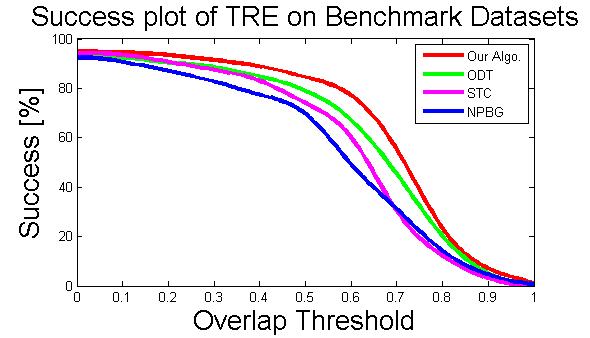}
\label{fig_gr}}
\hfil
\subfloat{\includegraphics[width=2in]{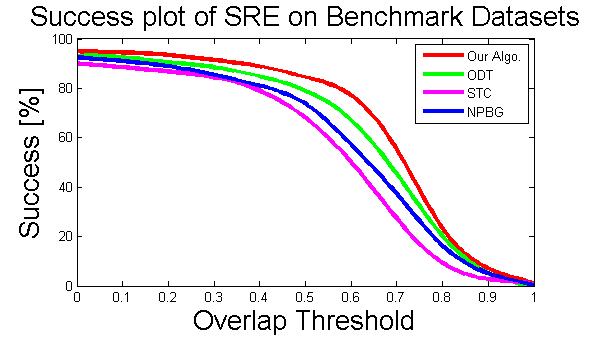}
\label{fig_gd}}
\hfil
\subfloat{\includegraphics[width=2in]{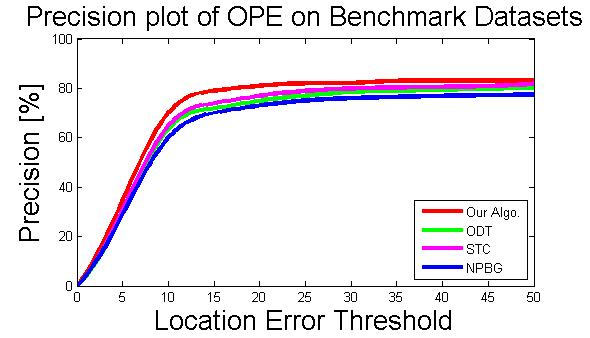}
\label{fig_gb}}
\hfil
\subfloat{\includegraphics[width=2in]{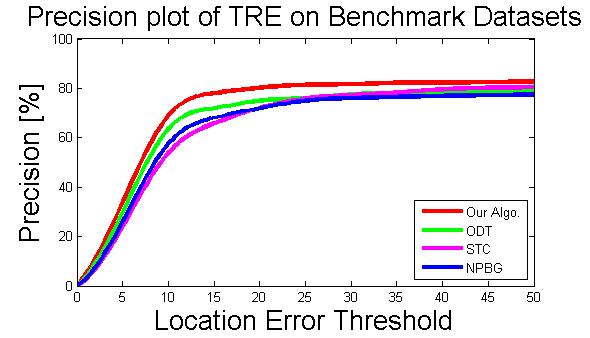}
\label{fig_gb1}}
\hfil
\subfloat{\includegraphics[width=2in]{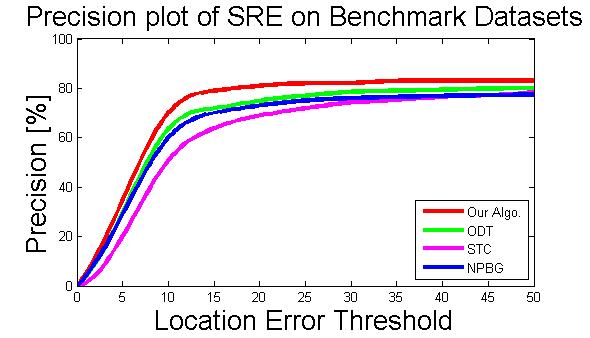}
\label{fig_gb1}}
\caption{Success and Precision plots for benchmark datasets (TB50 and TB100) using one pass evaluation (OPE), temporal robustness evaluation (TRE) and spatial robustness evaluation (SRE).}
\label{fig_7}
\end{figure*}

\begin{table*}[t]
\renewcommand{\arraystretch}{1.3}
\caption{Attribute-wise Experimental Results on Benchmark Datesets}
\label{table_er}
\begin{threeparttable}
\centering
\begin{tabular}{c||cccc|cccc|cccc}
\hline
\multirow{2}{*}{\textbf{Attribute}} &
\multicolumn{4}{c}{\textbf{Mean TD}} &
\multicolumn{4}{c}{\textbf{Mean FD}} &
\multicolumn{4}{c}{\textbf{Mean M}}\\

& \textbf{ODT}\cite{cite5:Hu} & \textbf{STC} \cite{cite17:Zhang} & \textbf{NPBG} \cite{cite24:Kim} & \textbf{Our Algo.} & \textbf{ODT}  & \textbf{STC}  & \textbf{NPBG}  & \textbf{Our Algo.} & \textbf{ODT}  & \textbf{STC}  & \textbf{NPBG}  & \textbf{Our Algo.}  \\
\hline\hline
{\textbf{OCC}} & 75\% & 73.77\% & {\color{blue}75.36}\% & {\color{red}81.1}\% & 4.79\% & {\color{blue}4.2}\% & 4.88\% & {\color{red}3.63}\% & {\color{blue}2.1}\% & 3.44\% & 3.77\% & {\color{red}1.47}\% \\
\hline
\textbf{ROT} & {\color{blue}83.63}\% & 81\% & 77.1\% & {\color{red}89.79}\% & 4.0\% & {\color{blue}2.1}\% & 2.2\% & {\color{red}2.0}\% & 3.97\% &  {\color{red}1.2}\% & {\color{blue}1.44}\% & {\color{red}1.2}\% \\
\hline 
\textbf{DEF} & {\color{blue}71.63}\% & 67\% & 67.1\% & {\color{red}73.33}\% & 4.75\% & 4.21\% &  {\color{red}3.86}\% & {\color{blue}3.94}\% & 3.37\% & {\color{blue}2.92}\% & 2.96\% & {\color{red}2.27}\% \\
\hline 
\textbf{BGC} & 61.45\% & {\color{blue}67}\%  & 57.1\% & {\color{red}67.73}\% & 5.9\% & 5.91\% & {\color{blue}5.77}\% & {\color{red}5.68}\% & {\color{blue}3.23}\% & 3.92\% & 3.96\% & {\color{red}3.0}\% \\
\hline
\end{tabular}
\begin{tablenotes}
      \item {\color{red}red}: rank1, {\color{blue}blue}: rank2      
\end{tablenotes}
\end{threeparttable}
\end{table*}

\begin{table}[t]
\renewcommand{\arraystretch}{1.3}
\caption{Attribute-wise Execution Time on the Benchmark Datasets}
\label{table_fps}
\begin{threeparttable}
\centering
\begin{tabular}{c c c c c}
\hline
\textbf{Attribute} & \textbf{ODT} \cite{cite5:Hu} & \textbf{STC} \cite{cite17:Zhang} & \textbf{NPBG} \cite{cite24:Kim} & \textbf{Our Algo.} \\
\hline\hline
 {\textbf{OCC}} & 55  & {\color{red}145}  & 47  & {\color{blue}75} \\								
\hline
\textbf{ROT}    & 56  & {\color{red}165}  & 49	 & {\color{blue}94} \\						    
\hline
\textbf{DEF}    & 50  & {\color{red}139}  & 41  & {\color{blue}82} \\					    
\hline
\textbf{BGC}    & 47  & {\color{red}117}  & 35  & {\color{blue}59} \\						    							
\hline
\end{tabular}
\begin{tablenotes}
      \item {\color{red}red}: rank1, {\color{blue}blue}: rank2      
\end{tablenotes}
\end{threeparttable}
\end{table}

\begin{table}[t]
\renewcommand{\arraystretch}{1.3}
\caption{Attribute-wise AUC/CLE Scores on the Benchmark Datasets}
\label{table_auccle}
\begin{threeparttable}
\centering
\begin{tabular}{c c c c c}
\hline
\textbf{Attribute} & \textbf{ODT} \cite{cite5:Hu} & \textbf{STC} \cite{cite17:Zhang} & \textbf{NPBG} \cite{cite24:Kim} & \textbf{Our Algo.} \\
\hline\hline
 {\textbf{OCC}} & 62.1/{\color{blue}6.79}    & {\color{blue}63.22}/7.33  & 61.4/10.49 & {\color{red}69.7}/{\color{red}4.9} \\								
\hline
\textbf{ROT}    & 65/5.5    	 & {\color{blue}67.83}/5.43   & 61.56/{\color{blue}5.0}		& {\color{red}70.77}/{\color{red}4.2} \\						    
\hline
\textbf{DEF}    & 61.19/15.22  & {\color{blue}62.89}/{\color{blue}11.97}  & 57.9/19.8  & {\color{red}68.44}/{\color{red}10.1} \\					    
\hline
\textbf{BGC}    & {\color{blue}58.3}/17.32  & 55.0/{\color{blue}15.56}   & 54.91/17.0  & {\color{red}65.75}/{\color{red}12.49} \\						    							
\hline
\end{tabular}
\begin{tablenotes}
      \item {\color{red}red}: rank1, {\color{blue}blue}: rank2      
\end{tablenotes}
\end{threeparttable}
\end{table}

\begin{figure*}
\centering
\subfloat{\includegraphics[width=3.5in]{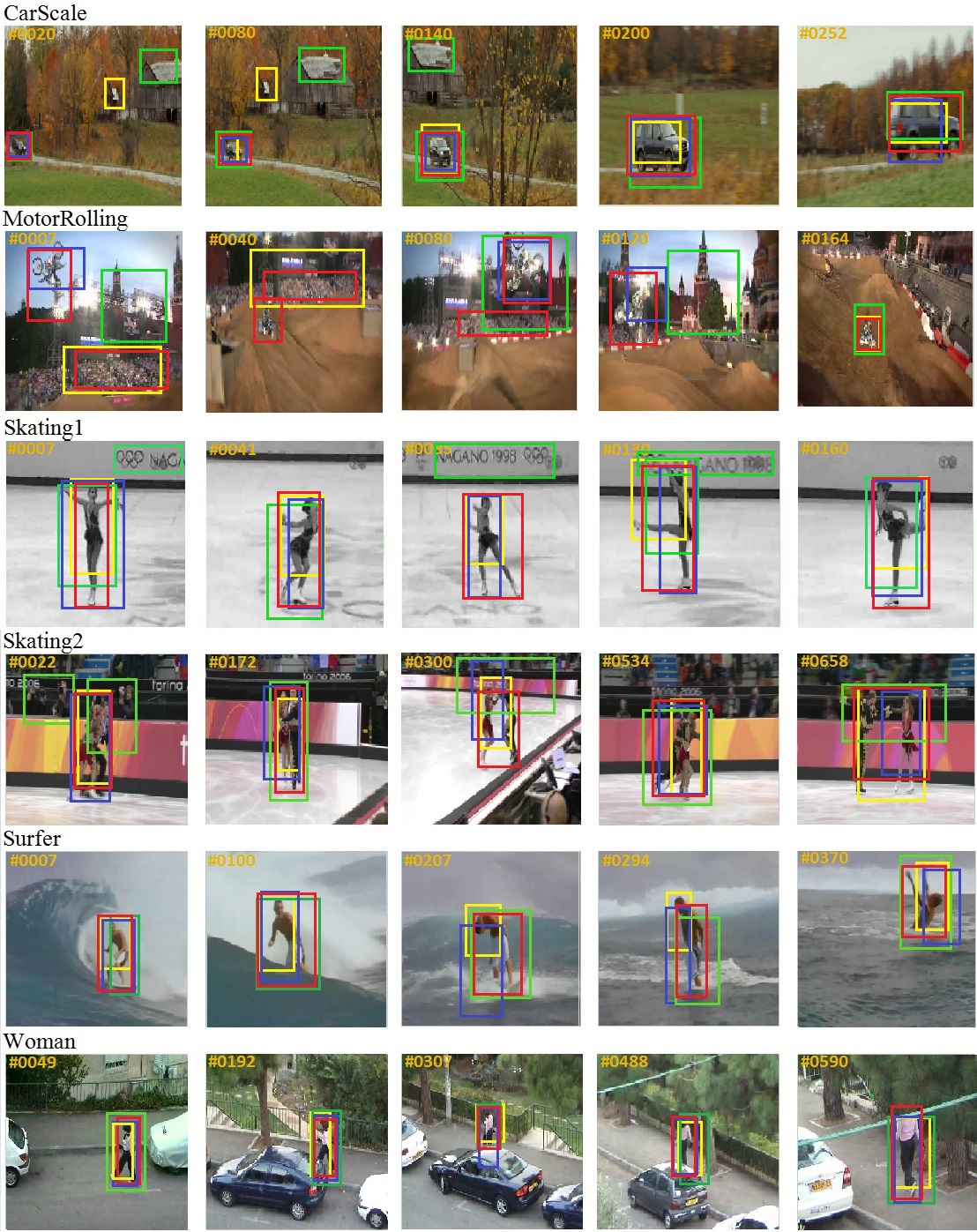}
\label{fig_go}}
\subfloat{\includegraphics[width=3.5in]{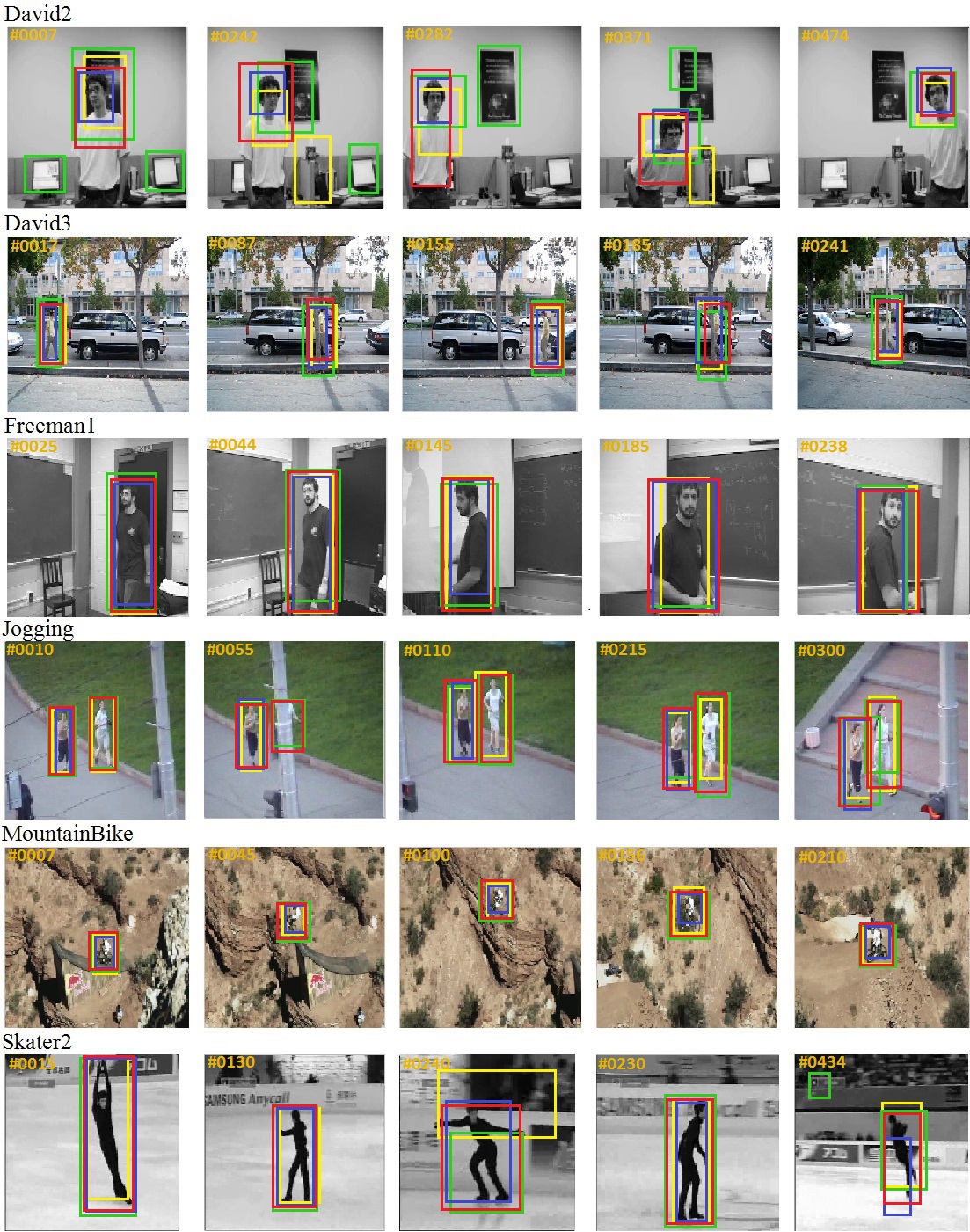}
\label{fig_gr}}
\hfil
\subfloat{\includegraphics[width=3.5in]{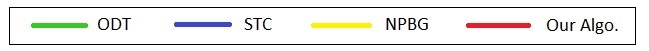}
\label{fig_lg}}
\caption{Results of Object detection and tracking by proposed algorithm and benchmark algorithms.}
\label{fig_8A}
\end{figure*}



We have implemented our algorithm in Matlab and conducted our experiments on a computer with an Intel Core i7 3.40 GHz CPU and 16 GB RAM. We have tested the performance of our algorithm on the benchmark dataset- TB-50 containing 50 sequences and TB-100, extension of TB-50 containing another 50 sequences \cite{cite3:Wu}. These datasets provide ground-truth markings and annotation with various attributes like occlusion, background clutter, rotation, illumination variation etc. on video sequences. Each video  contains one or two moving object/s in a moderate or very complex background. Depending on the major challenges present in environment and background of the videos, we have sorted them in following four attributes: 1) Occlusion (OCC)- 47 sequences, 2) both in-plane and out-of-plane rotation (ROT)- 71 sequences, 3) Deformation (DEF)- 41 sequences and 4) Background clutter including illumination variation (BGC)- 52 sequences. The significant property of our algorithm is that, we did not provide the initial state of object/s in the starting frame; object/s is automatically detected without initialization and training on sample data. We have also compared the effectiveness our algorithm with some of the state-of-the-art algorithms: moving object detection and tracking in videos with moving background (ODT) \cite{cite5:Hu}, tracking via Dense Spatio-Temporal Context Learning (STC) \cite{cite17:Zhang}, detection of moving objects with a moving camera using non-panoramic background model (NPBG) \cite{cite24:Kim}. 

\subsection{Experimental Setup}

To generate spatio-temporal blobs for each frame of each of the input image sequences we have used one single spatio-temporal Gabor filter bank. The base spatial frequency ($\omega$) for the filter bank is set as $1/4$ $cycles/pixel$. Three spatial orientations - $0^\circ,35^\circ,75^\circ$ are selected to calculate the center spatial frequencies ($\omega_{x_0}$'s, $\omega_{y_0}$'s) in horizontal and vertical directions as per (\ref{eqn_2}) and (\ref{eqn_3}). The center temporal frequencies ($\omega_{t_0}$'s) are tuned to $1/7,1/8,1/9$ $cycles/frame$. We have applied a heuristic method on our test set of input videos to select the optimal orientations and temporal frequencies. This design heuristic is achieved from domain specific information for a large class of problems. At first, filters calculated using orientations $0^\circ,45^\circ,90^\circ$ have been applied on test videos and it is observed that at orientation $0^\circ$ foreground is captured quite well whereas at other two orientations neither background nor foreground is prominently filtered. Then angles between $0^\circ,45^\circ$ are tested and at orientation $35^\circ$ desired result was achieved and in the same method orientation $75^\circ$ have been selected between $45^\circ,90^\circ$. Filters calculated from orientations of other quadrants are of similar nature and repeat the result obtained by the filters created from the angles of first quadrant. That is these filters have no significant contribution in foreground detection and as number of filters affects the speed of filtering process; three optimum orientations from the first quadrant have been decided. Same heuristic method is applied to select temporal frequencies. Lesser or higher frequencies than the selected frequencies have no significant effect in filtering the foreground.

The standard deviation of all the Gaussian envelopes for spatio-temporal complex sinusoids are set as $\sigma_x= \sigma_y=4$ and $\sigma_t=1$. Thus, the spatial extent of the kernel is twenty five (25) pixels in each direction and seven (7) frames is the temporal extent. Then for each image sequence, we have created three dimensional image blocks using seven (7) consecutive frames throughout the length of the image sequence. For example, the first image block of the video is created for $7^{th}$ frame and contains $1^{st}$ to $7^{th}$ frame, second image block is for $8^{th}$ frame and contains $2^{nd}$ to $8^{th}$ frame and so on. Other standard deviations for the Gaussian envelopes have also been applied on our test videos. It is observed that, spatial Gaussian envelopes of smaller standard deviation than the selected one; are inclined to emphasize minute localized details of a frame- lesser the size of spatial Gaussian envelopes, more localized details of the frame are captured. Thus if the intended object is large (occupies at least one-third of frame) then it may be detected by fragments appearing as many small objects and properties of  the actual objects (centroid, height and width, color etc.) are miscalculated which leads to erroneous tracking. Whereas, wider spatial Gaussian envelopes suppress the finer local structures in the frame. So, a small object (occupies less than one-fifth of the frame) may be retrieved in a diffused and distorted shape which also leads to erroneous object representation and performance of the tracking algorithm deteriorates. For medium sized objects, it will be detected with other spurious structures around it which again leads to erroneous calculation of features of object and unsuccessful tracking. If standard deviation of the temporal Gaussian envelope is decreased then number of frames in the spatio-temporal block is decreased, i.e. lesser number of frames are compared and analyzed at a time. Thus slower changes due to background variations will also be captured along with the faster changes due to movement of object. These unnecessary background information hinder the detection and tracking of actual object. On the other hand, if standard deviation of the temporal Gaussian envelope is increased then larger number of frames are compared and analyzed at a time. This leads to suppression of finer movements of moving object (like head rotation, joint movement etc.) which causes very fragmented and disconnected spatio-temporal blobs which are parts of the same object. Thus accuracy of detection of actual object and tracking decreases. If we measure the accuracy of object detection in 1 to 10 scale then Fig. \ref{fig_6} depicts the effect of increase or decrease of standard deviations of Gaussian envelopes on the accuracy of object detection.

\subsection{Evaluation and Analysis}

We have quantitatively evaluated and compared our algorithm using four parameters: Frames per second (FPS), True Detection (TD), False Detection (FD) and Missed Detection (MD). FPS is the count of annotated frames displayed per second. TD is calculated as the percentage of frames with successful object detection and tracking in each sequence : 
\begin{equation}
\label{eqn_17}
TD=\frac{n_{td}}{N}\times100
\end{equation}
where N = Total number of frames in an image sequence and $n_{td}$ = number of frames with truly detected object. We have measured the success of a frame as per the following rule- if in a frame, bounding box around a detected and tracked object overlaps with the bounding box of the ground truth i.e. $|C_t-C_g|\leq B_t/B_g$, where C's are the centroids and B's are the bounding boxes; then the frame is marked as a successful frame. If object is detected in a position of a frame where ground truth does not indicate any or detected bounding box does not overlap with the bounding box of the ground truth i.e. $|C_t-C_g|>B_t/B_g$; then the detection is considered as false detection and formulated as: 
\begin{equation}
\label{eqn_18}
FD=\frac{n_{fd}}{n_{td}+n_{fd}}\times100
\end{equation} If object is not detected in a frame, but ground truth value for that frame exists; then the situation is considered as Missed Detection and formulated as: 
\begin{equation}
\label{eqn_19}
MD=\frac{n_{md}}{n_{td}+n_{md}}\times100
\end{equation} 

We have executed our algorithm and other state-of-the-art algorithms on individual sequence of each of attributes (OCC, ROT, DEF and BGC) in TB50 and TB100 and calculated the above four metrics per sequence. For each tested algorithm, attribute-wise mean value of True Detection (TD), False Detection (FD) and Missed Detection (MD) is estimated and is summarized in Table \ref{table_er}. Attribute-wise mean of execution time for each evaluated algorithm is presented in Table \ref{table_fps}. Fig.\ref{fig_8A} contains the results of object detection and tracking in widely spaced sequential frames of some of the input image sequences. It is apparent from the results that, our proposed algorithm achieved best mean of true detection rate (Mean TD) for all the four attributes with respect to other state-of-the art algorithms. It also achieved lowest mean of false detection (Mean FD) and missed detection (Mean MD) for all the four attributes. As we have analyzed both spatial and temporal context of a frame simultaneously with respect to several previous consecutive frames; our algorithm is able to extract only significant spatio-temporal variation of a frame over background clutter, sudden object deformation or rotation or partial occlusion etc. On the other hand, both ODT \cite{cite5:Hu} and NPBG \cite{cite24:Kim} extracts spatial features from two consecutive frames \cite{cite5:Hu} or from current frame and a background model \cite{cite24:Kim}. Then spatial features are correlated temporally to register the images for object detection. So, these methods are susceptible to failure in detecting object due to erroneous feature extraction and tracking or highlighting insignificant context variation between background of two frames or sudden object deformation or rotation. NPBG performs better in latter phase as its background model becomes more precise with temporal update. Though STC \cite{cite17:Zhang} uses spatio-temporal context to detect and track object; this method learns significant spatio-temporal variations gradually and needs to be properly initialized manually or by some object detection algorithms. So, this method suffers heavily if object detection in first frame is not perfect and also suffers from lack of accuracy at the initial stage. Our algorithm is executed in an average of 81.5 frames per second (FPS) for all the sequences of benchmark datasets. We have achieved second best performance in terms of FPS for all four attributes after STC \cite{cite17:Zhang}. But, in STC only a smaller local context region is selected manually in first frame and is analyzed in subsequent frames to detect and track object. Whereas, in our method the whole frame is analyzed spatially and temporally to detect and track objects. Our method is faster than ODT \cite{cite5:Hu} and NPBG \cite{cite24:Kim}, as we have applied frequency domain analysis over spatial domain analysis to detect region of motion in each frame. 

We have also evaluated and compared our algorithm using one pass evaluation (OPE), temporal robustness evaluation (TRE) and spatial robustness evaluation (SRE) metrics \cite{cite3:Wu} and reported the results through precision curves and success plots in Fig. \ref{fig_7}. Precision curve expresses the percentage of frames in which center location error (CLE) of tracked object and ground truth centroids is within a given threshold and success plot expresses the percentage of successful frames as the tracked object and ground truth overlap score varies from 0 to 1. Given the bounding boxes of target object ($r_t$) and ground truth ($r_g$), the overlap score (S) is defined as: 
\begin{equation}
\label{eqn_20}
S=\frac{|r_t\cap r_g|}{|r_t\cup r_g|}
\end{equation} where, $\cap$ and $\cup$ represent the intersection and union operators respectively and $|\cdot|$ is the number of pixels in a region. One pass evaluation (OPE) is the usual test method whereas in temporal robustness evaluation (TRE), tracking algorithm is evaluated by starting at different frames and in spatial robustness evaluation (SRE), an algorithm is started with different bounding boxes. The TRE and SRE scores in Fig. \ref{fig_7} are the average of evaluations at different temporal and spatial initializations respectively. Our algorithm achieved 87.71\% average success rate and 80.73\% average precision rate for all the three evaluation metrics with respect to the usual overlap threshold ($t_0=0.5$) and CLE threshold ($p_0=20$) \cite{cite3:Wu}. As our method does not depend on spatial initialization and is able to detect moving object by spatio-temporal analysis, it performs well in SRE. It is also successful and precise in TRE, as our method depends on short-term temporal information of a frame and no long-term temporal learning is necessary as NPBG or STC. Our algorithm can start detecting moving objects from any frame in a sequence except first six frames. As STC requires long-term temporal learning, starting the algorithm at different frames diminishes its performance than OPE. Also, as STC is highly dependent on spatial initialization it suffers heavily due to improper spatial initialization. We also presented the attribute-wise area under curve (AUC) and average center location error (CLE) scores of all the four algorithms using one pass evaluation (OPE) on whole benchmark datasets in Table \ref{table_auccle}. The proposed algorithm exhibits best results with respect to the state-of-the-art methods in handling the challenges of all the four attributes.

\section{Conclusion And Future Work}
The proposed algorithm is successful in detecting actual moving objects in variable background of an image sequence without using any background information and additional sensor data. The idea of analyzing an image sequence spatio-temporally using three dimensional Gabor filter, is very effective for extracting spatio-temporal blobs as region of interest. The algorithm has successfully merged releted blobs into actual moving object by efficient implementation of Minimum Spanning Tree method. The algorithm is also successful in implementing Linear Assignment Problem efficiently for tracking an object and kalman filter to handle the occlusion. By the experiment so far it is observed that, the objects that are visible (prominently present) through the length of the video or a significant duration of the video; have been selected automatically as the actual moving object/s by local spatio-temporal analysis of the image sequence. It is also observed that occlusion for short duration does not affect the generation of spatio-temporal blobs. However, a group of objects moving with similar speed in a very close proximity is considered as a single object. 
The major advantage of our method is, it does not require initialization of object region at first frame or training on sample data to perform. We have achieved satisfactory results on benchmark videos in TB50 and TB100 and also the performance of our method is comparable and superior with respect to state-of-the-art methods \cite{cite5:Hu}, \cite{cite17:Zhang}, \cite{cite24:Kim}. In future, we would like to extend our work to detect and track object/s in extremely complex variable background like very crowded scene and/or presence of extreme illumination variation or other extreme challenges.


%

\appendices

\floatstyle{boxed} 
\restylefloat{figure}

\section{}
Algorithm to Update feature vector of each track is $Algorithm 1$, which is depicted in Fig. \ref{fig_alg1}.
\begin{figure*}
\caption{Algorithm to Update feature vector of each track }
\label{fig_alg1}
\begin{algorithmic}
\State \textbf{Algorithm 1:} Updating feature vector of tracks  
\\ \State \textbf{Input: } 
\State $[objID,cost]_{K\times L}$- cost matrix for the current frame
\State $Kalman_{1 \times K}$ - kalman prediction corresponding each track 
\State \textbf{Output: } Updated feature vector of track for the current frame
\begin{enumerate} 
\For{$k=$ each $Track_k$}
\ \ \If{$cost_{mn}==min(cost_{k1}...cost_{kM})$}
\ \ \ \ \State update features of $Track_k$ and $Kalman_k$ by the features of $Object_m$;
\ \ \ \ \State remove entry from cost matrix corresponding to $Object_m$
\ \ \EndIf
\EndFor
\ \ \State Label each track with null/no\_match entry in cost matrix as unassigned;
\ \ \State Label each object with entry in cost matrix as candidateNew;
\For{$i=$ each unassigned track}
\ \ \ \State $Track_i$ is updated by $Kalman_i$
\EndFor
\For{$j=$ each candidateNew}
\ \ \If{$min(cost_{k1}...cost_{kM})==\phi$} 
\ \ \ \ \State assign $Object_j$ to a new track;
\ \ \ElsIf{$min(cost_{k1}...cost_{kM})<\phi$} 
\ \ \ \ \State $Object_j$ is discarded; \Comment{as it may be an erroneous detection}
\ \ \EndIf
\EndFor 
\end{enumerate}
\State $Return$ updated $[Centroid_k, BB_k, GH_k]$ and $Kalman_k$ for each track
\end{algorithmic}
\end{figure*}


\ifCLASSOPTIONcaptionsoff
  \newpage
\fi



\bibliographystyle{IEEEtran}
%

%

\begin{IEEEbiography}[{\includegraphics[width=1in,height=1.25in,clip,keepaspectratio]{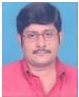}}]{Kumar S. Ray}
, PhD, is a Professor in the Electronics and Communication Sciences Unit at Indian Statistical Institute, Kolkata, India. He is an alumnus of University of Bradford, UK. Prof. Ray was a member of task force committee of the Government of India, Department of Electronics (DoE/MIT), for the application of AI in power plants. He is the founder member of Indian Society for Fuzzy Mathematics and Information Processing (ISFUMIP) and member of Indian Unit for Pattern Recognition and Artificial Intelligence (IUPRAI). 
His current research interests include artificial intelligence, computer vision, commonsense reasoning, soft computing, non-monotonic deductive database systems, and DNA computing. He is the author of two research monographs viz, Soft Computing Approach to Pattern Classification and Object Recognition, a unified conept, Springer, Newyork, and Polygonal Approximation and Scale-Space Analysis of closed digital curves, Apple Academic Press, Canada, 2013.
\end{IEEEbiography}
\begin{IEEEbiography}[{\includegraphics[width=1in,height=1.25in,clip,keepaspectratio]{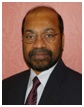}}]{Vijayan K. Asari}
, PhD, is a Professor in electrical and computer engineering and Ohio Research Scholars endowed chair in wide area surveillance at the University of Dayton, Dayton, Ohio, USA. He is the director of the Center of Excellence for Computer Vision and Wide Area Surveillance Research (UD Vision Lab) at UD. He received his bachelor’s degree in electronics and communication engineering from the University of Kerala, India, 1978, M Tech and PhD degrees in electrical engineering from the IIT, Madras in 1984 and 1994, respectively. He holds three patents and has published more than 500 research papers in the areas of image processing, computer vision, machine learning, pattern recognition, and high-performance digital architectures. 
\end{IEEEbiography}
\begin{IEEEbiography}[{\includegraphics[width=1in,height=1.25in,clip,keepaspectratio]{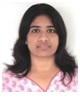}}]{Soma Chakraborty}
is working as a project linked scientist at Electronics and communication Sciences Unit of Indian Statistical Institute, Kolkata, India. Prior to joining ISI in December 2013, she worked in software industry as developer in healthcare and manufacturing domain for 7.8 years. She received her bachelor’s degree in Information Technology from West Bengal University of Technology, India in 2005 and MS degree in Software System from BITS, Pilani, India in 2012. Her research interest includes video image processing, computer vision and pattern recognition.
\end{IEEEbiography}




\end{document}